\documentclass[pdflatex,sn-mathphys-num]{sn-jnl}

\usepackage{graphicx}%
\usepackage{multirow}%
\usepackage{amsmath,amssymb,amsfonts}%
\usepackage{amsthm}%
\usepackage{mathrsfs}%
\usepackage[title]{appendix}%
\usepackage{xcolor}%
\usepackage{textcomp}%
\usepackage{manyfoot}%
\usepackage{booktabs}%
\usepackage{algorithm}%
\usepackage{algorithmicx}%
\usepackage{algpseudocode}%
\usepackage{listings}%

\usepackage{booktabs}
\usepackage{float}
\usepackage{colortbl}
\usepackage{bm}
\usepackage{wrapfig}
\usepackage{pifont}

\def\eg{\emph{e.g.}}

\def\ie{\emph{i.e.}}

\definecolor{hl}{RGB}{235,230,222}
\def\red#1{\textcolor{red}{#1}}
\def\blue#1{\textcolor{blue}{#1}}
\def\green#1{\textcolor{green}{#1}}
\def\violet#1{\textcolor{violet}{#1}}
\def\orange#1{\textcolor{orange}{#1}}

\def\cyan#1{\textcolor{cyan}{#1}}

\theoremstyle{thmstyleone}%
\theoremstyle{thmstyletwo}%

\theoremstyle{thmstylethree}%

\begin{document}

\title[Article Title]{	
Texture, Shape, Order, and Relation Matter: A New Transformer Design for Sequential DeepFake Detection}


\author[1]{\fnm{Yunfei} \sur{Li}}\email{liyunfei.x@gmail.com}

\author*[1,]{\fnm{Yuezun} \sur{Li}}\email{liyuezun@ouc.edu.cn}



\author[2]{\fnm{Baoyuan} \sur{Wu}}\email{wubaoyuan@cuhk.edu.cn}

\author[1]{\fnm{Junyu} \sur{Dong}}\email{dongjunyu@ouc.edu.cn}

\author[3]{\fnm{Guopu} \sur{Zhu}}\email{guopu.zhu@hit.edu.cn}

\author[4]{\fnm{Siwei} \sur{Lyu}}\email{siweilyu@bufflao.edu}

\affil[1]{\orgdiv{School of Computer Science and Technology}, \orgname{Ocean University of China}, \orgaddress{\city{Qingdao}, \country{China}}}


\affil[2]{\orgdiv{School of Data Science}, \orgname{The Chinese University of Hong Kong}, \orgaddress{\city{Shenzhen}, \country{China}}}

\affil[3]{\orgdiv{ School of Cyberspace Science}, \orgname{Harbin Institute of Technology}, \orgaddress{\city{Harbin}, \country{China}}}

\affil[4]{\orgname{University at Buffalo, SUNY}, \orgaddress{\city{New York}, \country{US}}}


\abstract{Sequential DeepFake detection is an emerging task that predicts the manipulation sequence in order. Existing methods typically formulate it as an image-to-sequence problem, employing conventional Transformer architectures. However, these methods lack dedicated design and consequently result in limited performance. As such, this paper describes a new Transformer design, called {TSOM}, by exploring three perspectives: Texture, Shape, and Order of Manipulations. Our method features four major improvements: \ding{182} we describe a new texture-aware branch that effectively captures subtle manipulation traces with a Diversiform Pixel Difference Attention module. \ding{183} Then we introduce a Multi-source Cross-attention module to seek deep correlations among spatial and sequential features, enabling effective modeling of complex manipulation traces. \ding{184} To further enhance the cross-attention, we describe a Shape-guided Gaussian mapping strategy, providing initial priors of the manipulation shape. \ding{185} Finally, observing that the subsequent manipulation in a sequence may influence traces left in the preceding one, we intriguingly invert the prediction order from forward to backward, leading to notable gains as expected. Building upon TSOM, we introduce an extended method, {TSOM++}, which additionally explores Relation of manipulations: \ding{186} we propose a new sequential contrastive learning scheme to capture relationships between various manipulation types in sequence, further enhancing the detection of manipulation traces. We conduct extensive experiments in comparison with several state-of-the-art methods, demonstrating the superiority of our method. The code has been released at \url{https://github.com/OUC-VAS/TSOM}.}




\maketitle

\section{Introduction}\label{sec:intro}

Recent advancements in deep generative techniques have significantly improved the visual quality of generated faces. One such technique, DeepFake, allows for effortless manipulation of faces, producing results that are often imperceptible to the naked eye. This technique has found applications in virtual reality and movie special effects, but it also raises severe security concerns~\cite{shao2025deepfake,xu2024learning, yang_ijcai21}. To address these concerns, various forensic methods have been proposed to expose DeepFakes, garnering considerable attention~\cite{li2019exposing,yang2019exposing,li2018ictu,yin2024fine,yu2024mining,xu2024fd}. 

\begin{figure*}[!t]
    \centering
    \includegraphics[width=\linewidth]{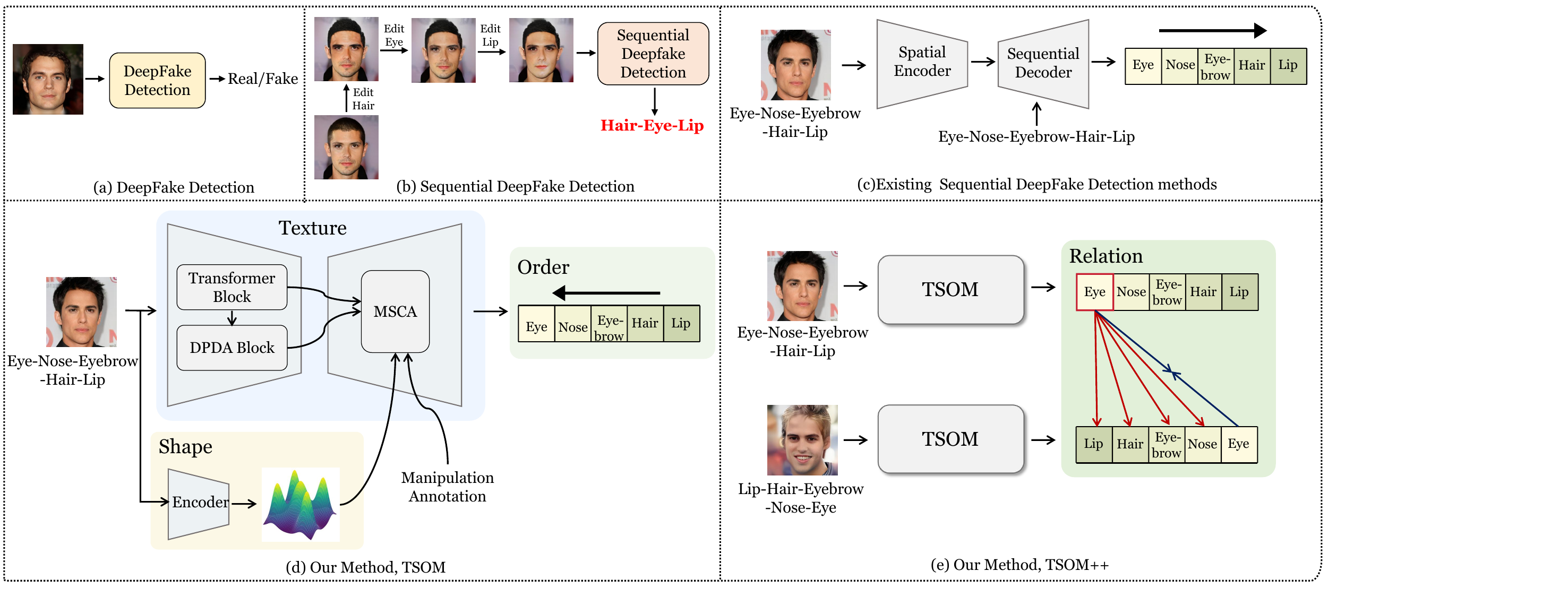}
    \vspace{-0.6cm}
    \caption{\small \cyan{(a)} and \red{(b)} correspond to One-step DeepFake Detection (ODD) and sequential DeepFake Detection (SDD). \violet{(c)} illustrates the process of typical Sequential DeepFake Detection (SDD) architecture. \orange{(d)} represents our proposed method TSOM. In contrast to (c), our method focuses on exploring three perspectives: Texture, Shape and Order of Manipulations, and introduces four major improvements: a Diversiform Pixel Difference Attention (DPDA) module, a Multi-source Cross-attention (MSCA) module, a Shape-guided Gaussian Mapping (SGM) strategy, an Inverted Order Prediction (IOP). \green{(e)} represents our method TSOM++, which introduces a Sequential Manipulations Contrastive Learning (SMCL).}
    \label{fig:comp}
\end{figure*}

Conventional forensic methods mainly focus on determining whether a given face image is authentic or forged, which is typically framed as a binary classification task (see Fig.~\ref{fig:comp}\cyan{(a)}). However, with the rapid evolution of face editing tools, the manipulation processes have become more versatile and accessible, enabling the use of multiple specific operations iteratively until the desired effects are reached. In this context, existing methods can only identify the final step of manipulation, providing limited insights into the manipulation details. To overcome this limitation, a new forensics task called \textbf{Sequential DeepFake Detection (SDD)}~\cite{shao2022seqdeepfake} has emerged recently. In contrast to the conventional DeepFake detection (\ie, One-step DeepFake Detection (ODD)), SDD attempts to uncover the sequences of face manipulations in order. For example, a DeepFake face may be created by sequentially editing regions such as the hair, eyes, and lips. Given such a DeepFake face, SDD can predict the manipulation sequence as ``Hair-Eye-Lip'' (see Fig.~\ref{fig:comp}\red{(b)}). This emerging task improves the practicality of DeepFake detection in real-world scenarios, offering more comprehensive evidence for traceability and forensic interpretability (\eg, pinpointing where, when, and what manipulations are applied). This endeavor expands the horizon of DeepFake detection and poses a fresh challenge to the field.

In general, the task of SDD can be conceptualized as an image-to-sequence problem~\cite{oriol2015cap}, where the typical solution involves leveraging Transformer-based architectures to predict the manipulation sequences~\cite{xia2024mmnet,shao2022seqdeepfake}. Concretely, the model utilizes a spatial encoder with conventional self-attention mechanisms to extract spatial manipulation traces. Then, a sequential decoder is employed to establish conventional cross-attention~\cite{vaswani2017attention} with the encoder features and manipulation annotation sequence, facilitating the capture of sequential manipulation traces (see Fig.~\ref{fig:comp}\violet{(c)}). 

\begin{figure}[!t]
    \centering
    \includegraphics[width=0.6\linewidth]{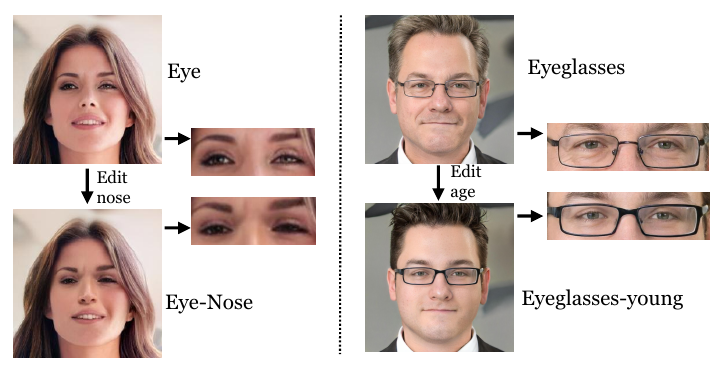}
    \vspace{-0.2cm}
    \caption{\small The impact of the sequential manipulation on facial attributes. The second manipulation impacts the first manipulation: (left) Eye and (right) Eyeglasses.}
    \label{fig:order}
\end{figure}

Despite these methods showing certain effectiveness, many critical perspectives are lacking and require further investigation: \textbf{1) Neglect of texture details:}
Generative models often exhibit imperfect semantic disentanglement, wherein one specific manipulation may affect the entire facial structure, leaving subtle traces on unintended manipulated regions. These subtle traces often manifest more prominently in the textures~\cite{liu2020global,yang2021mtd}. However, existing methods primarily use conventional self-attention, which emphasizes global correlations and tends to overlook fine-grained texture cues. 
\textbf{2) Absence of manipulation shape guidance:} Current pipelines detect manipulations holistically, ignoring their specific shapes. Note that the manipulation regions often exhibit various shapes. Incorporating shape information could help the network better localize and interpret manipulations. While existing method~\cite{shao2022seqdeepfake} incorporates the scale and position of manipulation regions, they could not effectively characterize irregular manipulation regions. 
\textbf{3) Suboptimal prediction order:} Existing methods predict the manipulation sequences in a forward order, the same as the actual manipulation order. However, later manipulations often obscure earlier ones (see Fig.~\ref{fig:order})), \ie, the preceding manipulation ``hides beneath'' the subsequent one, akin to the layers of an ``onion''. The first manipulation step (\eg, Hair as shown in Fig.~\ref{fig:comp}\red{(b)}) acts as the core of the onion. Extracting this core without peeling away its outer layers (\eg, Eye and lip as shown in Fig.~\ref{fig:comp}\red{(b)}) would be improper. 

Delving into these perspectives, we propose a new Transformer called \underline{\textbf{TSOM}} (\underline{Te}xture, \underline{S}hape and \underline{O}rder of \underline{M}anipulations) to expose sequential DeepFakes (see Fig.~\ref{fig:comp}\orange{(d)}). Our method introduces four major improvements: \ding{182} In the encoder, we develop two branches for capturing different types of manipulation traces. One branch, termed the vanilla branch, employs conventional Transformer blocks to capture global relationships. The other branch, a new texture-aware branch, is designed to capture subtle manipulation traces. Within this branch, we describe a Diversiform Pixel Difference Attention (DPDA) module, which dynamically integrates various texture extraction operations into the self-attention mechanism, guided by information from the vanilla branch. \ding{183} We further propose a new Multi-source Cross-attention (MSCA) module to effectively model correlations between spatial features from the encoder and sequential manipulation annotations. Unlike traditional cross-attention, our module processes multiple features simultaneously and employs a multi-level fusion strategy for more effective correlations. \ding{184} To further enhance cross-attention, we propose a Shape-guided Gaussian Mapping (SGM) strategy to provide priors on manipulation shapes. These priors are estimated using a Variational Auto-encoder (VAE)~\cite{vae2013}. \ding{185} During training, we invert the order of the manipulation sequences. This simple ``trick'' notably improves the detection performance, as expected. 

Based on TSOM, we describe \underline{\textbf{TSOM++}}, an extended method that further explores the relation of manipulations: \ding{186} We observe that the same manipulation types across different DeepFake instances (\eg, ``lip" edits) tend to share visual similarities, whereas different manipulation types, even within the same face, often differ markedly (See Fig.~\ref{fig:comp}\green{(e)}). By leveraging these relationships, the model can better understand manipulation semantics and improve overall performance. To facilitate this, we propose a sequential contrastive learning scheme to capture these relation patterns.   

Our methods are evaluated on a sequential DeepFake dataset~\cite{shao2022seqdeepfake} and compared with several state-of-the-art methods. The results demonstrate its superiority in detecting sequential DeepFakes. Furthermore, we conduct a comprehensive analysis of each component within our methods, corroborating the efficacy of the proposed strategies.

This paper extends our preliminary work presented at \texttt{WACV (oral) 2025}~\cite{li2025texture}, with the following major improvements: 1) We introduce TSOM++, an extended method that explores a fresh dimension -- the relations among manipulations. To effectively capture these relations, we describe a sequential contrastive learning scheme that encourages representations of the same manipulations to cluster closely, while separating those of different manipulations. 2) We extensively validate the capability of TSOM++ against a range of recent methods and conduct a comprehensive study on the effect of its components, providing valuable insights for future research.


\section{Related Works}

\smallskip
\noindent\textbf{One-step DeepFake Detection (ODD).}
To combat the threats posed by DeepFake, a considerable amount of methods have been proposed for DeepFake detection~\cite{shao2025deepfake, zhi2022region,yuting2023tall,shi2023impli,guo2023controllable,wang2023dynamic}. These methods can typically be classified into different categories in terms of the way of feature capturing. The methods of~\cite{li2018ictu,yang2019exposing,agarwal2019protecting,qi2020deeprhythm,yu2024benchmarking} expose DeepFake by capturing abnormal biological signals, such as eye blinking, behavior patterns, or heart rhythm. Other methods focus on extracting the spatial artifacts introduced either by face generation or the blending operations~\cite{li2019exposing,li2020face,zhao2021learning,shiohara2022detecting}. Instead of capturing the spatial artifacts, the methods of~\cite{li2024sa, liu2021spatialphase,dzanic2020fourier} demonstrate the effectiveness of frequency information and capture representative traces in the frequency domain. Moreover, many recent methods design dedicated models to capture the traces automatically~\cite{zhao2021multi,guo2023controllable,wang2023dynamic}. 
However, these methods can only predict authenticity, which can not handle and analyze the multi-step alterations in practice.

\smallskip
\noindent\textbf{Sequential DeepFake Detection (SDD).}
Sequential DeepFake detection is an emerging task focused on predicting the sequential manipulations in order. This task is typically formulated as the image-to-sequence problem, utilizing architectures tailored for sequential tasks in detection. The work of~\cite{shao2022seqdeepfake} is pioneering in addressing this problem, employing the Transformer architecture~\cite{vaswani2017attention}, which includes a conventional encoder and decoder with vanilla self-attention mechanisms.
The work of ~\cite{shao2025robust} extends~\cite{shao2022seqdeepfake} by establishing a stronger correspondence between image-sequence pairs through image-text contrastive learning and image-text matching, thereby enhancing the performance of the model.
Subsequently, MMNet~\cite{xia2024mmnet} improved its performance through the incorporation of multi-collaboration and multi-supervision modules. 
The work of ~\cite{hong2024contrastive} is the most recent method for the SDD task, which achieves top performance by decomposing the original task into two sub-tasks: multi-attribute deepfake detection and attribute ranking, and using the corresponding classification and ranking losses to constrain the model.
Despite their promising results, this task remains largely unexplored.

\section{Method}
Our method TSOM is developed on a Transformer-based architecture comprising a spatial encoder and a sequential decoder. The encoder is designed to extract spatial manipulation features, including a vanilla branch and a texture-aware branch. The vanilla branch employs several conventional Transformer blocks to capture the global correlations. In the texture-aware branch, we introduce the \textbf{Diversiform Pixel Difference Attention (DPDA)} module to capture subtle manipulation traces (Sec.~\ref{sec:dpda}). 
The decoder models the sequential relation based on the spatial features and corresponding sequence annotations to predict the facial manipulation sequence. In this decoder, we propose a \textbf{Multi-source Cross-attention (MSCA)} strategy to fuse the extracted spatial features with the sequence annotation embeddings, enabling effective modeling of sequential relations (Sec.~\ref{sec:MSCA}). Moreover, we introduce \textbf{Shape-guided Gaussian Mapping (SGM)} to enhance the effect of cross-attention in sequential modeling (Sec.~\ref{sec:shape}). We then introduce the procedure for \textbf{Inverted Order Prediction (IOP)} (Sec.~\ref{sec:invert}). 
Subsequently, we describe the extended method, TSOM++, which enhances TSOM by exploring the relation among sequential manipulations achieved by a newly proposed \textbf{Sequential Manipulations Contrastive Learning (SMCL)} scheme (Sec.~\ref{sec:smcl}). Finally, we outline the overall loss functions (Sec.~\ref{sec:loss}).

\subsection{Texture-aware Branch}
\label{sec:dpda}
Denote a face image as $\bm{\mathcal{I}} \in \mathbb{R}^{H \times W \times 3}$. We first tokenize this face image using a simple CNN stem $\mathcal{F}$ following~\cite{He2015} as $\mathbf{x}^{0} = \mathcal{F}(\bm{\mathcal{I}}) + \mathbf{p}, \mathbf{x}^{0} \in \mathbb{R}^{h \times w \times c}$, where $\mathcal{F}(\bm{\mathcal{I}})$ indicates the feature maps from CNN stem and $\mathbf{p}$ is a positional embedding. These tokens are then flattened in spatial dimension as the input of the vanilla branch and texture-aware branch in the spatial encoder. 

To extract these subtle manipulation traces, we develop a new Transformer Block for extracting texture features. Our block draws inspiration from central difference operators~\cite{yu2020searching,juefei-xu2017lbcnn}, which are convolutions used for detecting local textures. In our method, we derive and extend their essence from the convolution style to the self-attention mechanism, and propose a Diversiform Pixel Difference Attention (DPDA) module that captures fine-grained spatial traces while retaining the global view within the vanilla self-attention.

\begin{figure}[!t]
    \centering
    \includegraphics[width=0.6\linewidth]{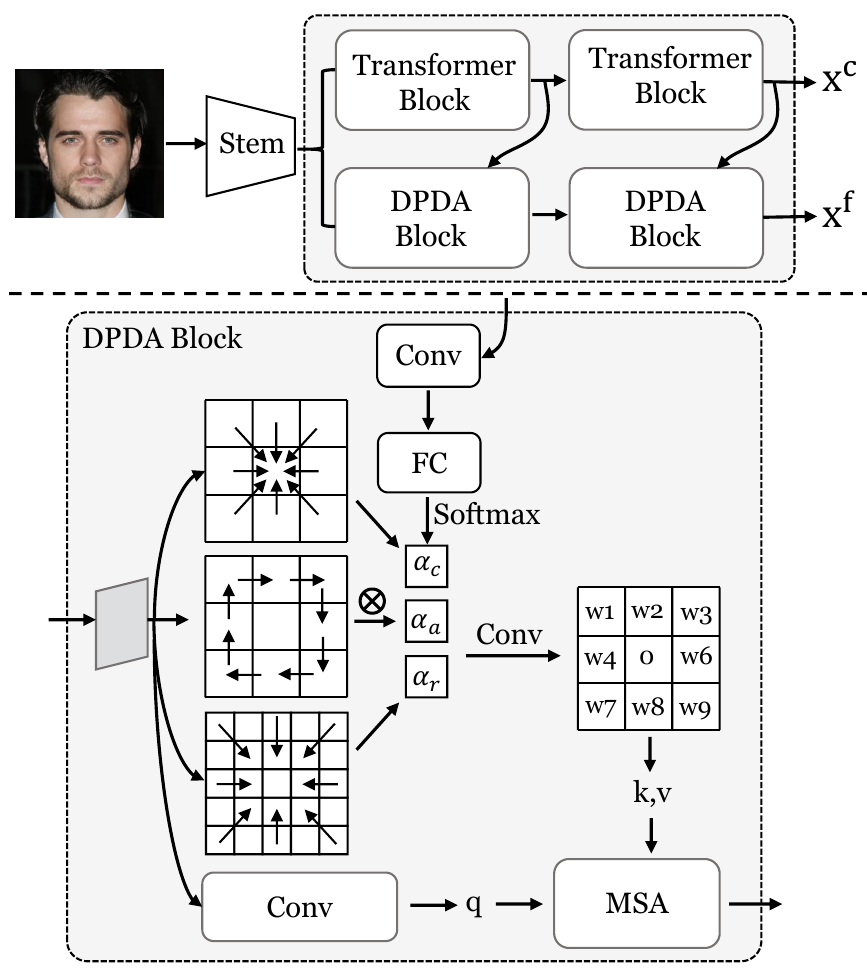}
    \vspace{-0.3cm}
    \caption{\small Overview of the spatial encoder (top). Illustration of DPDA module (bottom). MSA denotes Multi-head Self-attention.}
    \label{fig:encoder}
\end{figure}

\smallskip
\noindent\textbf{Revisit of Central Difference Convolution.}
Central difference convolution is a special case of convolution, which emphasizes the center-oriented gradients in the local receptive field region. Given a feature map $\mathbf{x}$, this operation can be formulated as 
\begin{equation}
\small
\begin{aligned}
    \mathbf{x}'(p_0) = & \sum_{p_n \in \mathcal{O}} \mathbf{w}(p_n) (\mathbf{x}(p_0+p_n) - \mathbf{x}(p_0)) \\
\end{aligned}
\end{equation}
where $\mathbf{w}$ is the convolution kernel, $\mathcal{O}$ denotes the local receptive field, $p_0$ indicates the current location of input feature map $\mathbf{x}$ and output feature map $\mathbf{x}'$, and $p_n$ is the relative position in $\mathcal{O}$.

\smallskip
\noindent\textbf{Diversiform Pixel Difference Attention (DPDA).}
Inspired by the central difference operations, we extend these operations and integrate such spirit into the self-attention mechanism following~\cite{yu2020searching,yu2020fas,miao2023f}. 
Specifically, given the feature map $\mathbf{x}$, we first perform a vanilla convolution on it to generate the query feature, which is defined as
\begin{equation}
\small
\begin{aligned}
    \mathbf{x}^{Q}(p_0) = & \sum_{p_n \in \mathcal{O}} \mathbf{w}(p_n) \mathbf{x}(p_0+p_n)    
\end{aligned}
\end{equation}
Then we would like to capture the fine-grained spatial features as the source of key and value features. Since the central difference operations focus on the center-oriented gradients, they may overlook other texture patterns, which hinders the effectiveness of capturing manipulation traces.  
To obtain a diverse range of gradient information, we leverage the concept of Extended Local Binary Pattern (ELBP), which can encode the pixel relations from various perspectives, including central, angular, and radial directions~\cite{zhuo2021pixel}. The angular direction calculates the difference between adjacent pixels in the outline of the kernel in a certain order, while the radial direction calculates the difference between radial adjacent pixels, see Fig.~\ref{fig:encoder}. To obtain the key and value features, we adaptively ensemble these operations using a learnable coefficient vector, which is formulated as 
\begin{equation}
\small
\begin{aligned}
    \mathbf{x}^{K,V}(p_0) = \sum_{p_n \in \mathcal{O}} \mathbf{w}(p_n)
    \left (
    \underbrace{\left [\alpha_c,\alpha_a,\alpha_r \right ]}_{\bm{\alpha}} 
    \left [
    \begin{array}{l}
        \mathcal{D}_c \\
        \mathcal{D}_a \\
        \mathcal{D}_r
    \end{array}
    \right ] \right)
\end{aligned}
\end{equation}
In this equation, $\mathcal{D}_c, \mathcal{D}_a, \mathcal{D}_r$ denote central, angular, and radial difference operations respectively, which can be defined as
\begin{equation}
\small
\begin{aligned}
    \mathcal{D}_c & = \mathbf{x}(p_0+p_n) - \mathbf{x}(p_0), \\
    \mathcal{D}_a & = \mathbf{x}(p_0+p_n) - \mathbf{x}(p_{n-1}), \\
    \mathcal{D}_r & = \mathbf{x}(p_0+p_n) - \mathbf{x}(p_{\phi(n)})
\end{aligned}
\end{equation}
We use $3 \times 3$ as the local receptive field $\mathcal{O}$ for instance and set $\mathcal{O} = \{(-1, -1), (-1, 0),...,(1,1) \}$. Thus $\phi(n)$ in radial difference operation can defined as $\phi(n) = (2x,2y)$ if $n = (x,y)$. $\bm{\alpha}$ is the coefficient vector to balance the contribution of each operation.

\smallskip
\noindent\textbf{Adaptive Balance.} Since various DeepFake faces exhibit diverse manipulation traces, the coefficient vector $\bm{\alpha}$ should be different. Thus, we design a strategy to dynamically regulate the influence of different DPDA operations. As the vanilla branch concentrates on global correlations, it has deeper insight into the manipulation composition. Hence, we set $\bm{\alpha}$ trainable and adapt it based on the knowledge from the vanilla branch. Let $\mathbf{x}'$ be the intermediate feature of a block within the vanilla branch. We develop a sub-network $\mathcal{H}$ to learn $\bm{\alpha}$ as
\begin{equation}
\small
\bm{\alpha} = {\rm Softmax} ( \mathcal{H}(\mathbf{x}'))
\end{equation}
After obtaining $\mathbf{x}^{Q}$ and $\mathbf{x}^{K,V}$, we project the query feature $\mathbf{x}^{Q}$ into query tokens, and project $\mathbf{x}^{K,V}$ into key and value tokens. Within each head, ${Q}_i = \mathbf{x}^{Q} {W}^{Q}_i, {K}_i = \mathbf{x}^{K,V} {W}^{K}_i, {V}_i = \mathbf{x}^{K,V} {W}^{V}_i$. Then the multi-head self-attention is calculated using 
\begin{equation}
\small
    \begin{aligned}
    & h_i = {\rm Softmax}(Q_i K^{\top}_i / \sqrt{d_i})V_i, \\
    & {\rm MSA}(\mathbf{x}) = {\rm Concat}(h_1,...,h_n) W^{o},
    \end{aligned}
    \label{eq:msa}
\end{equation}
where $h_i$ is the self-attention of $i$-th head. Then we concatenate all these heads and perform a linear transformation with $W^{o}$ to obtain the final output.

\begin{figure*}[!ht]
    \centering
    \includegraphics[width=0.7\linewidth]{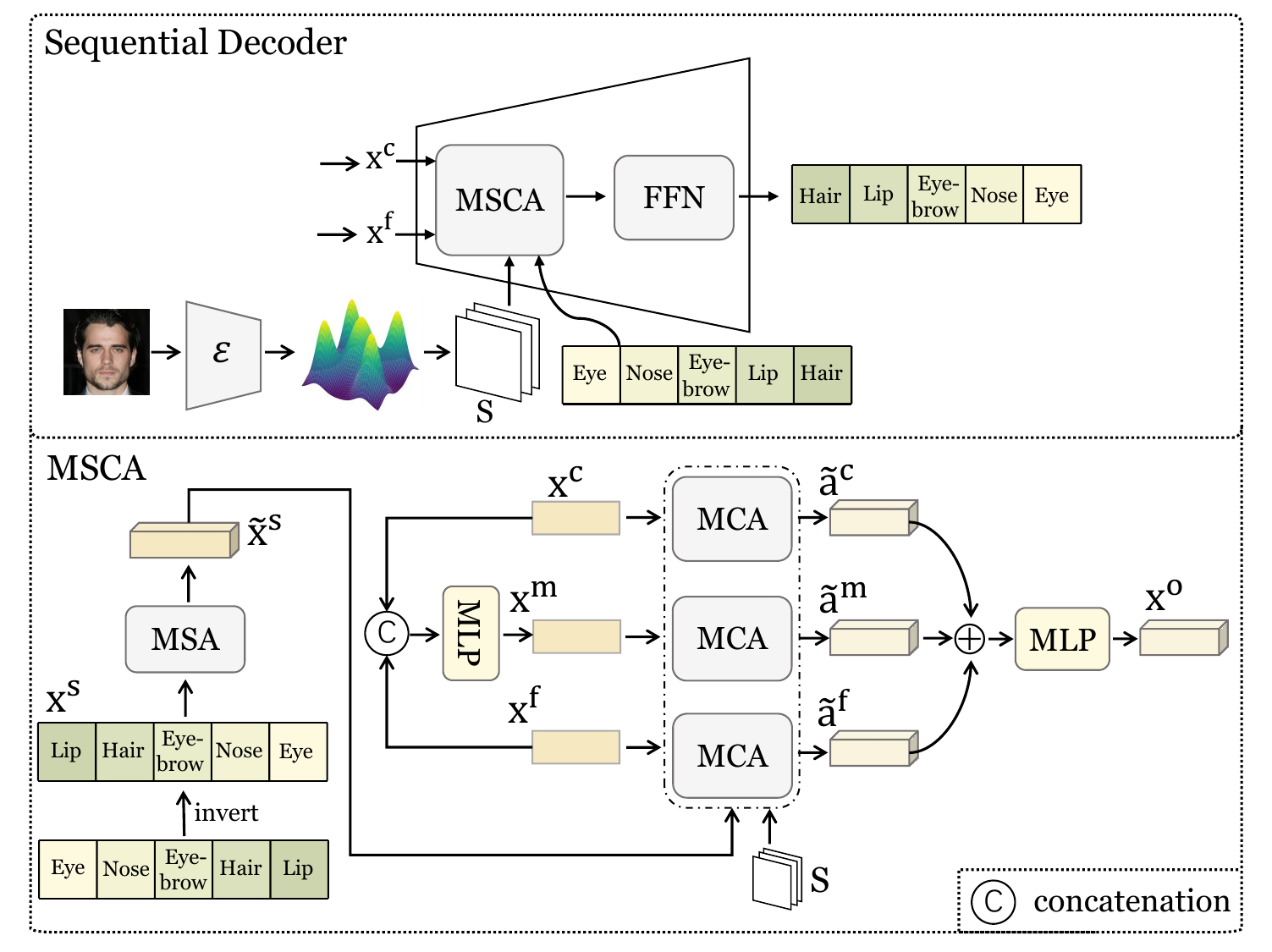}
    \vspace{-0.2cm}
    \caption{\small Overview of the sequential decoder (left). Illustration of Multi-source Cross-attention(MSCA). MCA denotes Multi-head Cross-attention.}
    \label{fig:decoder}
\end{figure*}

\subsection{Multi-source Cross-attention (MSCA)}
\label{sec:MSCA}
\smallskip
\noindent\textbf{Revisit of Conventional Cross-attention.}
Typically, the cross-attention~\cite{vaswani2017attention} is employed for modeling such correlation~\cite{shao2022seqdeepfake}. However, these mechanisms usually process the features from only two sources $\mathbf{x}_{1}$ and $\mathbf{x}_{2}$ (\eg, the encoder features and sequential embedding). The conventional Multi-head Cross-attention (MCA) can be defined as
\begin{equation}
\small
    \begin{aligned}
        & Q_i = \mathbf{x}_{1} W^{Q}_i, K_i = \mathbf{x}_{2} W^{K}_i, V_i = \mathbf{x}_{2} W^{V}_i, \\
        & h_i= {\rm Softmax}(Q_i K^{\top}_i / \sqrt{d_i})V_i,\\
        & {\rm MCA}(\mathbf{x}_{1},\mathbf{x}_{2}, \mathbf{x}_{2})=  {\rm Concat}(h_1,...,h_n) W^{o}
    \end{aligned}
\end{equation}

\smallskip
\noindent\textbf{Operations of MSCA.} Let $\mathbf{x}^{c}$ and $\mathbf{x}^{f}$ represent the features from the coarse-grained and fine-grained branches. Let $\mathbf{x}^{s}$ be the embedding of inverted sequential manipulation annotations. In our task, it is intuitive to perform separate MCA operations on $\mathbf{x}^{s}$ with $\mathbf{x}^{c}$ and $\mathbf{x}^{f}$. However, this integration fails to explore deeper connections between these three features. As such, we propose a {Multi-source Cross-attention (MSCA)} module to capture the correlations among multiple sources. The overview of the entire module is illustrated in Fig.~\ref{fig:decoder}.

Firstly, we concatenate coarse-grained features $\mathbf{x}^{c}$ and fine-grained features $\mathbf{x}^{f}$ along the channel dimension. Subsequently, we facilitate interactions between the two different levels of fabricated features throuth an MLP structure, yielding shallowly fused sequencetial features $\mathbf{x}^{m}$. Then, by performing self-attention computations on $\mathbf{x}^{s}$, we extract features that exploit correlations among annotations of sequence tampering. These operations can be formulated as follows:
\begin{equation}
\small
\mathbf{x}^{m} = {\rm MLP}({\rm Concat}( \mathbf{x}^{c}, \mathbf{x}^{f})), \;
\tilde{\mathbf{x}}^{s} = {\rm MSA}(\mathbf{x}^{s})
\end{equation}
Next, we implement a second-stage fusion strategy, performing multi-head cross-attention between $\tilde{\mathbf{x}}^{s}$ and $\mathbf{x}^{c}$, $\mathbf{x}^{m}$, $\mathbf{x}^{f}$ respectively, as:
\begin{equation}
\small
    \begin{aligned}
        \tilde{\mathbf{a}}^{c} & = {\rm MCA}(\tilde{\mathbf{x}}^{s}, \mathbf{x}^{c}, \mathbf{x}^{c}), \mathbf{a}^{c} = {\rm LN}(\tilde{\mathbf{a}}^{c} + \tilde{\mathbf{x}}^{s}), \\
        \tilde{\mathbf{a}}^{m} & = {\rm MCA}(\tilde{\mathbf{x}}^{s}, \mathbf{x}^{m}, \mathbf{x}^{m}), \mathbf{a}^{m} = {\rm LN}(\tilde{\mathbf{a}}^{m} + \tilde{\mathbf{x}}^{s}), \\
        \tilde{\mathbf{a}}^{f} & = {\rm MCA}(\tilde{\mathbf{x}}^{s}, \mathbf{x}^{f}, \mathbf{x}^{f}), \mathbf{a}^{f} = {\rm LN}(\tilde{\mathbf{a}}^{f} + \tilde{\mathbf{x}}^{s})
    \end{aligned}
\end{equation}
Finally, we directly sum up the three types of features and then pass them through an MLP to filter out irrelevant noise, thereby uncovering identical manipulation patterns and obtaining the final sequential deepfake features, following the operations as
\begin{equation}
\small
\mathbf{A} = \tilde{\mathbf{a}}^{c} + \tilde{\mathbf{a}}^{m} + \tilde{\mathbf{a}}^{f}, \;
\tilde{\mathbf{x}}^{o} = {\rm MLP}(A), \;
\mathbf{x}^{o} = {\rm LN}(\tilde{\mathbf{x}}^{o} + \mathbf{A})
\end{equation}

\begin{figure}[!t]
    \centering
    \includegraphics[width=0.6\linewidth]{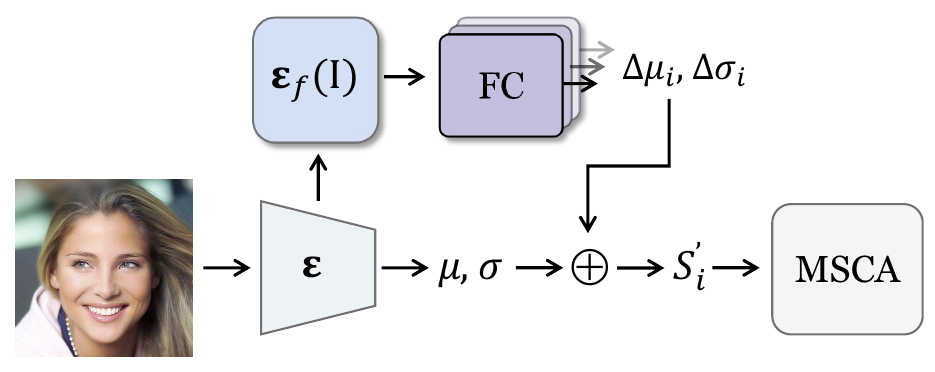}
    \vspace{-0.2cm}
    \caption{\small Integration with MSCA.}
    \label{fig:shape}
\end{figure}

\subsection{Shape-guided Gaussian Mapping}
\label{sec:shape}
We propose a Shape-guided Gaussian Mapping (SGM) strategy to leverage the strong spatial position priors associated with each manipulation component in the faces. Since each manipulation has its specific characteristic, its manipulation shape likely follows a certain distribution. By learning these distributions, we can predict the probability of each position on a face image being manipulated. This probability can serve as shape priors to further enhance the effectiveness of MSCA.

\smallskip
\noindent\textbf{Distribution Estimation.}
Since no ground truth is provided, estimating these distributions is non-trivial.
Inspired by unsupervised learning strategies, we construct a Variational Auto-encoder (VAE)~\cite{vae2013} to estimate these distributions. Specifically, we construct an encoder $\mathcal{E}$, which transforms the input image $\bm{\mathcal{I}}$ into means $\bm{\mu}$ and standards $\bm{\sigma}$ of these distributions as $\bm{\mu} ,\bm{\sigma} = \mathcal{E}(\bm{\mathcal{I}})$. Note that $\bm{\mu} ,\bm{\sigma}$ are vectors containing mean and standard of each manipulation.
To ensure these distributions align with the input image, we employ a decoder $\mathcal{D}$ for reconstruction based only on the estimated distributions.

\smallskip
\noindent\textbf{Integration with MSCA.}
We create a grid map $M$ that is composed of the $x$- and $y$- coordinates of each position. This grid map is then used to predict the probability of each position in different facial manipulations. This process can be formulated as $S = \mathcal{N}(M \; | \; \bm{\mu}, \bm{\sigma})$,
where $S$ is the probability map for each spatial position. 
Considering different heads in MSCA have different focuses. To properly integrate SGM with MSCA, we craft different probability maps $\{S'_i\}_{i=1}^{h}$ for each head. Specifically, we use $\bm{\mu},\bm{\sigma}$ as the base of all heads and predict the head-specific offset $\Delta \bm{\mu}_{i},\Delta \bm{\sigma}_{i}$ for $i$-th head. These offsets are obtained as $\Delta \bm{\mu}_{i},\Delta \bm{\sigma}_{i}={\rm FC}(\mathcal{E}_{f}(\bm{\mathcal{I}}))$, where $\mathcal{E}_{f}(\bm{\mathcal{I}})$ denotes the last feature of encoder $\mathcal{E}$. Therefore, the probability map $S'_i$ for $i$-th head can be defined as 
\begin{equation}
\small
S'_{i} = \mathcal{N}(M \; | \; \bm{\mu}+\Delta \bm{\mu_{i}} ,\bm{\sigma}+\Delta \bm{\sigma_{i}})
\end{equation}
This process is illustrated in Fig.~\ref{fig:shape}. Then these maps are integrated into ${\rm MCA}(\tilde{\mathbf{x}}^{s},\mathbf{x}^{c},\mathbf{x}^{c})$ and ${\rm MCA}(\tilde{\mathbf{x}}^{s},\mathbf{x}^{f},\mathbf{x}^{f})$ operations in MSCA, by calculating the attention as
\begin{equation}
\small
h_i= {\rm Softmax}(Q_i K^{\top}_i / \sqrt{d_i} + \eta S'_{i})V_i
\end{equation}

\subsection{Inverted Order Prediction}
\label{sec:invert}
In the training phase, we invert the annotation of the manipulation sequence. For instance, as shown in Fig.~\ref{fig:decoder}, the manipulated face is annotated as ``eye-nose-eyebrow-hair-lip'', we rearrange it to ``lip-hair-eyebrow-nose-eye" as the input of the MSCA module for sequential modeling. This strategy enables the model to first predict the final manipulation (\eg, hair) and understand the consequent disturbance caused by it. Then the model concentrates on the preceding manipulation (\eg, lips). 
The rationale behind this strategy is that the subsequent manipulation (\eg, noise) may affect the preceding ones (\eg, eye), making it more challenging to detect them (recall Fig.~\ref{fig:order}). In contrast, preceding manipulation (\eg, hair) is not influenced by subsequent ones, making it more easily detectable.
In summary, this strategy offers the advantage that, compared to forward sequence predictions, each step in the inverted sequence predictions is only influenced by itself and remains independent of other manipulations.

\begin{figure}
    \centering
    \includegraphics[width=0.8\linewidth]{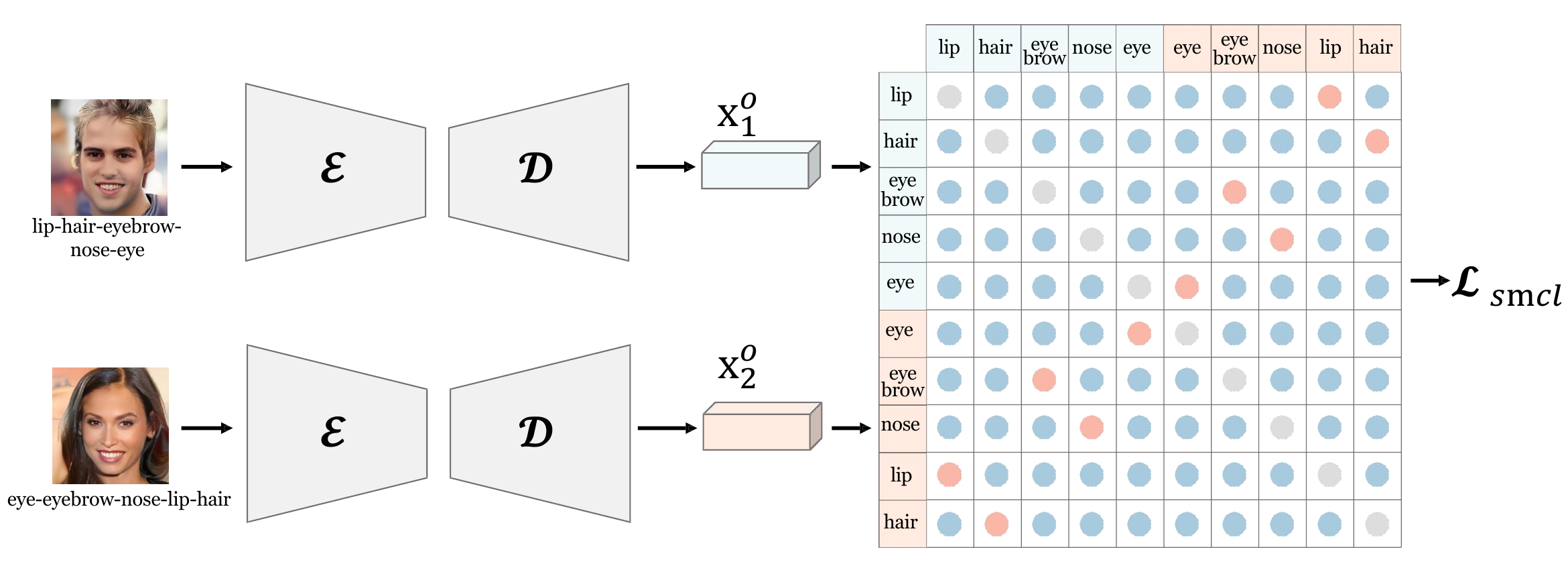}
    \caption{The Pipeline of the Sequential Manipulation Contrastive Learning (SMCL). Red circles indicate pairs to be pulled closer, blue circles indicate pairs to be pushed apart, and gray circles represent pairs that should be ignored.}
    \label{fig:scl}
\end{figure}

\subsection{Sequential Manipulations Contrastive Learning}\label{sec:smcl}
To further enhance the model's understanding of manipulation patterns, we describe TSOM++ by introducing a new Sequential Manipulations Contrastive Learning (SMCL) scheme, which captures intra- and inter-sequence relationships among manipulations. The overall SMCL pipeline is illustrated in Fig.~\ref{fig:scl}.

Consider two Deepfake face samples: one annotated with``lip-hair-eyebrow-nose-eye" and the other with ``eye-eyebrow-nose-lip-hair". We categorize the manipulation relation into three types: 1) \textit{Same manipulation in different faces}. For instance, if ``lip" is manipulated in both faces, their corresponding features should be similar. 2) \textit{Different manipulations within same face}. For example, the ``lip" and ``hair" manipulations within a single face should yield distinct feature representations. 3) \textit{Different manipulations across faces}. For example, the representation of the ``lip" manipulation in one face should differ from the ``eye" manipulation in another.

To depict these relations, we describe the SMCL framework. Denote $\mathbf{x}^o_1$ and $\mathbf{x}^o_2$ as the features of two faces after MSCA module. Each feature has the dimension of $n \times m$, where $n$ is the maximum length of manipulations and $m$ corresponds to the size of the manipulation ``vocabulary'' (\eg,  ``lip'', ``eye'', etc). If a face has fewer than $n$ manipulations, the remaining positions are padded with a special token \texttt{NM} (\ie, no manipulation). Assuming a batch contains two faces, the SMCL loss can be expressed as
\begin{equation}
    \mathcal{L}_{\rm smcl} = - \log \frac{
    \exp\left( \delta({f}^c, {g}^c)/\tau \right)}
    {\exp\left( \delta({f}^c, {g}^c)/\tau \right) + 
    \sum\limits_{d \neq c} \exp\left( \delta({f}^c, {f}^d)/\tau + \delta({f}^c, {g}^d)/\tau + \delta({g}^c, {g}^d)/\tau \right)},
\end{equation}
where $f$ and $g$ denote feature elements from $\mathbf{x}^o_1$ and $\mathbf{x}^o_2$, and $c,d$ represent the manipulation category of $f$ (\eg, c is ``lip'', d is ``eye''). Here, 
$\tau$ is a temperature hyper-parameter and $\delta$ is the similarity metric.
This contrastive setup encourages the model to learn nuanced relationships among manipulations by bringing similar manipulations closer and pushing dissimilar ones apart. Fig.~\ref{fig:scl} illustrates the relation matrix. The SMCL can pull close the pairs with red circles and push away the pairs with blue circles.

\subsection{Objectives}
\label{sec:loss}

In the training phase, the auto-regressive mechanism~\cite{alex2013} is employed to model sequence relationships with the cross-entropy loss $L_{\rm ce}$. 
This loss measures the error of the predicted manipulation sequence and ground truth.
To learn the shape priors, we perform the image reconstruction loss $L_{\rm rec}$ and KLD loss $L_{\rm kld}$ on constructed auto-encoder $(\mathcal{E},\mathcal{D})$ as in~\cite{vae2013}. Then the SMCL loss term is added into the overall objectives, which can be written as 
\begin{equation}
\small
    \begin{aligned}
    L = L_{\rm ce}+ \lambda _{1}L_{\rm rec} + \lambda _{2}L_{\rm kld} + \lambda _{3} L_{\rm smcl},
\end{aligned}
\end{equation}
where ${\lambda}_1,{\lambda}_2,{\lambda}_3$ are weights for balancing these loss terms.

\section{Experiments}
\subsection{Experimental Settings}
\label{sec:set}

\smallskip
\noindent\textbf{Datasets.} To the best of our knowledge, there is only one public dataset for sequential DeepFake detection that proposed in~\cite{shao2022seqdeepfake}. This dataset contains two types of sequential DeepFake manipulations, the facial
components manipulation and the facial attributes manipulation. The first track comprises $35,166$ face images that are subjected to replacing operations on specific facial components. There are $28$ manipulation types in this track, with a maximum sequence length of five. The second track contains $49,920$ face images with $26$ types of attribute manipulations. The maximum length of each manipulation is also five. Its split of training, validating, and testing set is $8:1:1$.

\smallskip
\noindent\textbf{Evaluation Metrics.}
In line with previous work~\cite{shao2022seqdeepfake}, we employ Fixed Accuracy (FACC) and Adaptive Accuracy (AACC) for evaluation. To calculate the FACC score, the sequence annotations shorter than five are padded with the label ``no manipulation'' and then compare the predictions with the padded sequence annotation. On the other hand, due to the nature of the auto-regress mechanism, the model can stop predicting upon encountering the End-of-Sequence (EOS) symbol, resulting in various lengths of the predicted sequence. The AACC score is calculated by comparing the predictions with the sequence annotations in terms of dynamic lengths.

\smallskip
\noindent\textbf{Implementation Details.}
Each branch in the spatial encoder has two Transformer blocks. Each block contains four attention heads. The sequential decoder consists of two Transformer blocks. In the training phase, the image size is set to $256 \times 256$ with augmentations such as horizontal Flipping and adjusting the brightness of the image. We employ the AdamW optimizer with a weight decay of $10^{-4}$. The initial learning rates are set to $10^{-4}$ for CNN stem and $2e^{-5}$ for Transformer architecture. We utilize a warm-up strategy of $20$ epochs, with a batch size of $32$. The total epochs are set to $170$, with a $10\%$ reduction in learning rate every $50$ epochs. The parameters in the loss function are set as: $\lambda_1 = 0.1, \lambda_2 = 0.01, \lambda_3 = 0.01$.

\begin{table*}[!ht]
\centering
\small
\caption{\small Performance of different methods on detecting sequential DeepFakes.}
\vspace{-0.2cm}
\resizebox{\linewidth}{!}{
\begin{tabular}{>{\raggedright}p{5cm}|p{1.2cm}<{\centering}|p{1.2cm}<{\centering}|p{1.2cm}<{\centering}|p{1.2cm}<{\centering}|p{1.2cm}<{\centering}|p{1.2cm}<{\centering}}
\hline
\multirow{2}{*}{Methods}          & \multicolumn{2}{c|}{Facial-components}  & \multicolumn{2}{c|}{Facial-attributes}    & \multicolumn{2}{c}{Average} \\ \cline{2-7}
                                    & FACC  & AACC  & FACC  & AACC  & FACC  & AACC  \\  \hline
Multi-Cls  (ResNet-34)~\cite{wang2021canmulti}            & 69.66      & 50.54        & 66.99     & 46.68   & 68.33   & 48.61  \\
Multi-Cls   (ResNet-50)~\cite{wang2021canmulti}           & 69.65      & 50.57        & 66.66     & 46.00   & 68.16   & 48.29    \\
DETR  (ResNet-34)~\cite{nicolas2020detr}             & 69.87      & 50.63        & 67.93     & 48.15  & 68.90   & 49.39      \\
DETR  (ResNet-50)~\cite{nicolas2020detr}             & 69.75      & 49.84        & 67.62     & 47.99   & 68.69   & 48.92     \\
DRN~\cite{sheng2019drn}                  & 66.06      & 45.79        & 64.42     & 43.20   & 65.24   & 44.50     \\
DRN$^*$~\cite{sheng2019drn}             & 69.38        & 49.76        & 67.66        & 47.31        & 68.52   & 48.54
\\
MA~\cite{hanqing2021ma}                   & 71.31      & 52.94        & 67.58     & 47.48    & 69.45   & 50.21    \\
MA$^*$~\cite{hanqing2021ma}          & 70.12        & 51.82        & 68.91        & 49.19        & 69.52        & 50.51
\\
Two-Stream~\cite{yuchen2021two-stream}              & 71.92      & 53.89        & 66.77     & 46.38   & 69.35 & 50.14     \\
Two-Stream$^*$~\cite{yuchen2021two-stream}   & 72.00        & 53.97     & 65.20   & 44.39    & 68.60      & 49.18     \\
SeqFake-Former (ResNet-34)~\cite{shao2022seqdeepfake}  & 72.13      & 54.80        & 67.99     & 48.32   & 70.06  & 51.56    \\ 
SeqFake-Former (ResNet-34)$^*$~\cite{shao2022seqdeepfake} & 70.97  & 52.65 & 68.40     & 48.93   & 69.69   & 50.79  \\
SeqFake-Former (ResNet-50)~\cite{shao2022seqdeepfake}  & 72.65      & 55.30        & 68.86     & 49.63   & 70.76   & 52.47    \\ 
SeqFake-Former (ResNet-50)$^*$~\cite{shao2022seqdeepfake} & 69.12   & 50.03     & 67.93     & 48.65    & 68.53  & 49.34 \\
MMNet (ResNet-50)~\cite{xia2024mmnet}            & 73.93      & 56.83        & 69.27      & 50.44   & 71.60   & 53.64   \\

MILNet(ResNet-50) ~\cite{hong2024contrastive}   &74.54  & -     &69.58  & - & 72.06     & -\\
MILNet(Swin-Transformer) ~\cite{hong2024contrastive}  & 74.97 & -   &70.02  & -  & 72.50    & -\\

\rowcolor{hl}\textbf{TSOM} (ResNet-34)            & 75.21      & 59.74        & 69.18      & 50.54   & 72.20  & 55.14   \\ 
\rowcolor{hl}\textbf{TSOM++}(ResNet-34)     & 75.61     & \textbf{60.21}     & 69.62 & 50.97
&72.62  & 55.59 \\

\rowcolor{hl}\textbf{TSOM} (ResNet-50)            & 75.53      & 59.67        & 69.80     & 51.21   & 72.67   & 55.44    \\
\rowcolor{hl}\textbf{TSOM++}(ResNet-50) & \textbf{75.94}     & 60.16     & \textbf{70.48} & \textbf{51.63}
&\textbf{73.21}  & \textbf{55.90} \\ \hline
\end{tabular}
}
\label{tab:main}
\end{table*}

\subsection{Results}
Table~\ref{tab:main} compares the performance of our method with other counterparts, including Multi-Cls, DETR~\cite{nicolas2020detr}, DRN~\cite{sheng2019drn}, MA~\cite{hanqing2021ma}, Two-Stream~\cite{yuchen2021two-stream}, SeqFake-Former~\cite{shao2022seqdeepfake} respectively. Multi-Cls and DETR serve as baseline methods adapted from models for general vision tasks.
Multi-Cls performs multi-label classification by directly classifying the manipulated images into multiple classes. DETR is a modified object detector that replaces object queries with manipulation annotation. DRN, MA, and Two-Stream are DeepFake detection methods adapted by substituting the binary classification head with a multi-classification head. SeqFake-Former~\cite{shao2022seqdeepfake} is a dedicated method for sequential DeepFake detection. MMNet~\cite{xia2024mmnet} incorporates the multi-collaboration module and multi-supervision strategy to improve the performance of sequential deepfake detection. MILNet is the latest work achieving the state-of-the-art performance. The results of Multi-Cls, DETR, DRN, MA, and Two-stream are referred from~\cite{shao2022seqdeepfake}. \textbf{Note that only DRN, MA, Two-stream, and SeqFake-Former have released their codes.} For better evaluation, we reproduce them with their released codes (marked by $*$).
As shown in Table~\ref{tab:main}, our method TSOM outperforms others in both tracks, averaging $72.67\%$ and $55.44\%$ in FACC and AACC scores on ResNet-50. Compared to SeqFake-Former, our method improves by $2.88\%,4.37\%$ and $0.94\%,1.58\%$ in these two tracks on ResNet-50. Moreover, our method averagely surpasses MMNet by $1.07\%$ and $1.8\%$ in FACC and AACC. 
 Compared to original TSOM, the extended TSOM++ achieves superior performance, improving FACC and AACC by {0.40\%, 0.47\%, 0.44\%, and 0.43\%} with ResNet-34, and 0.41\%, 0.49\%, 0.68\%, and 0.42\% with ResNet-50, on both datasets. Moreover, despite relying only on basic architectures such as ResNet, our TSOM++ method exceeds the most recent MILNet~\cite{hong2024contrastive} by 1.4\% and 0.9\% in FACC score on both datasets, even though MILNet employs a more advanced Swin Transformer model\footnote{Note that MILNet has not released the codes, and only FACC score are reported in original paper.}. These results demonstrate the effectiveness of our methods.


\begin{table*}[! t]
\centering
\small
\caption{\small Performance of different methods on challenging scenarios.}
\vspace{-0.2cm}
\resizebox{1\linewidth}{!}{
\begin{tabular}{>{\raggedright}p{4.5cm}|p{1.2cm}<{\centering}|p{1.2cm}<{\centering}|p{1.2cm}<{\centering}|p{1.2cm}<{\centering}|p{1.2cm}<{\centering}|p{1.2cm}<{\centering}}
\hline
\multirow{2}{*}{Methods}          & \multicolumn{2}{c|}{Facial-components$^\dagger$}     & \multicolumn{2}{c|}{Facial-attributes$^\dagger$}     & \multicolumn{2}{c}{Average} \\ \cline {2-7} 
& FACC     & AACC      & FACC      & AACC    & FACC    & AACC    \\  \hline
Two-Stream                  & 24.00      & 12.74        & 31.55     & 16.92      & 27.78    & 14.83 \\
DRN                   & 48.11     & 14.11        & 50.35     & 20.62     & 49.23    &  17.37  \\
MA              & 59.08      & 33.38        & 67.20     & 47.07      & 63.14    & 40.23   \\
SeqFake-Former (ResNet-34)  & 58.70      & 31.43        & 62.88     & 41.88    & 60.79   & 36.66   \\ 
SeqFake-Former (ResNet-50)  & 57.35      & 29.50        & 64.81     & 44.49    & 61.08  & 37.00   \\ 
\rowcolor{hl}\textbf{TSOM} (ResNet-34)            & 61.79      & 36.80        & 67.44      & 47.14  & 64.62  & 41.97  \\  
\rowcolor{hl}\textbf{TSOM++} (ResNet-34)            & 62.21      & 36.99        & 67.85      & 47.30  & 65.03  & 42.15  \\  
\rowcolor{hl}\textbf{TSOM} (ResNet-50)            & 62.52      & 37.21        & 67.33      & 48.03  & 64.93  & 42.62  \\  

\rowcolor{hl}\textbf{TSOM++} (ResNet-50)            & \textbf{63.08}      & \textbf{37.62}        & \textbf{67.96}      & \textbf{48.36}  & \textbf{65.52}  & \textbf{42.99}  \\  \hline
\end{tabular}
}
\label{tab:perturbations}
\end{table*}

\smallskip
\noindent\textbf{In Challenging Scenarios.}
In real-world scenarios, images often undergo various post-processing operations. To these scenarios, we perform post-processing operations on the facial-components and facial-attributes datasets. Specifically, for each image, we randomly select one or two processes from a set that includes adding Gaussian noises, shifting RGB channels, compression, converting to gray, and jittering colors. This increases the difficulty of original datasets, which we refer to as Facial-components$^\dagger$ and Facial-attributes$^\dagger$. We then evaluate different methods under these challenging scenarios. As shown in Table \ref{tab:perturbations}, while all methods suffer a notable performance drop, our model still outperforms others, demonstrating its robustness against external disturbances.
{Notbly, our TSOM++ further improves the average FACC and AACC scores by 0.42\%,0.18\% with ResNet-34, and by 0.59\%,0.37\% with ResNet-50 on both datasets, demonstrating the effectiveness of the proposed Sequential Manipulation Contrastive Learning (SMCL), even under challenging scenarios. }

\begin{table*}[!ht]
    \small
    \centering
	\vspace{-0.2em}
	\caption{\small Effect of each component on Facial-components (top) and Facial-attributes (bottom) datasets.}
 \vspace{-0.2cm}
 \resizebox{\linewidth}{!}{
		\begin{tabular}{l|c|c|c|c|c|c}
			\hline
			\multicolumn{1}{c|}{\multirow{2}{*}{Setting}}     & \multicolumn{2}{c|}{Resnet-34}    & \multicolumn{2}{c|}{Resnet-50}    & \multicolumn{2}{c}{Average}\\ \cline {2-7}
			& FACC          & AACC           & FACC         & AACC    & FACC    & AACC  \\ \hline
			BL                          & 70.45  \blue{(+0.00)}                    & 52.18    \blue{(+0.00)}                   & 69.47    \blue{(+0.00)}               & 50.89  \blue{(+0.00)}    & 69.96 \blue{(+0.00)}    & 51.34 \blue{(+0.00)}  \\ 
                BL+DPDA                             & 71.07  \blue{(+0.62)}                   & 52.84 \blue{(+0.66)}                      & 71.06  \blue{(+1.59)}                 & 52.80 \blue{(+1.91)}       & 71.07 \blue{(+1.11)}   & 52.82 \blue{(+1.48)}    \\ 
                BL+DPDA+MSCA                      & 71.74  \blue{(+1.29)}                   & 54.05  \blue{(+1.87)}                    & 72.04 \blue{(+2.57)}                 & 54.42  \blue{(+3.53)}      & 71.89 \blue{(+1.93)}    & 54.24 \blue{(+2.90)}  \\ 
                BL+DPDA+MSCA+IOP                   & 74.22  \blue{(+3.77)}                    & 58.19  \blue{(+6.01)}                     & 74.30  \blue{(+4.83)}                 & 58.05  \blue{(+7.16)}       & 74.26 \blue{(+4.30)} & 58.12 \blue{(+6.78)}  \\ 
                \rowcolor{hl}BL+DPDA+MSCA+IOP+SGM                          & 75.21 \blue{(+4.76)}                     & 59.74  \blue{(+7.56)}                     & 75.53  \blue{(+6.06)}                 & 59.67 \blue{(+8.78)}  
                & 75.37 \blue{(+5.41)}   & 59.71 \blue{(+8.37)}\\   
                \rowcolor{hl}BL+DPDA+MSCA+IOP+SGM+SMCL & \textbf{75.61} \blue{(+5.16)}      & \textbf{60.21}\blue{(+8.03)}        & \textbf{75.94}\blue{(+6.47)}      & \textbf{60.16}\blue{(+9.27)}  & \textbf{75.78}\blue{(+5.82)}  & \textbf{60.19}\blue{(+8.85)}  \\  \hline

                BL                          & 66.61  \blue{(+0.00)}                    & 46.47    \blue{(+0.00)}                   & 65.82    \blue{(+0.00)}               & 45.52  \blue{(+0.00)}    & 66.22 \blue{(+0.00)}    & 46.00 \blue{(+0.00)}  \\ 
                BL+DPDA                             & 66.78  \blue{(+0.17)}                   & 46.65 \blue{(+0.18)}                      & 66.10  \blue{(+0.28)}                 & 45.86 \blue{(+0.34)}       & 66.44 \blue{(+0.22)}   & 46.26 \blue{(+0.26)}    \\ 
                BL+DPDA+MSCA                      & 67.06  \blue{(+0.45)}                   & 47.11  \blue{(+0.64)}                    & 66.32 \blue{(+0.50)}                 & 45.88  \blue{(+0.36)}      & 66.69 \blue{(+0.47)}    & 46.50 \blue{(+0.50)}  \\ 
                BL+DPDA+MSCA+IOP                   & 67.69  \blue{(+1.08)}                    & 48.12  \blue{(+1.65)}                     & 66.56 \blue{(+0.74)}                 & 46.50  \blue{(+0.98)}       & 67.13 \blue{(+0.91)} & 47.31 \blue{(+1.31)}  \\ 
                \rowcolor{hl}BL+DPDA+MSCA+IOP+SGM      & 69.18 \blue{(+2.57)} & 50.54 \blue{(+4.07)}                     & 69.80  \blue{(+3.98)}                     & 51.21  \blue{(+5.69)}                 & 69.49 \blue{(+3.27)}  
                & \textbf{50.88} \blue{(+4.88)}   \\   
                \rowcolor{hl}BL+DPDA+MSCA+IOP+SGM+SMCL & \textbf{69.62} \blue{(+3.01)}    & \textbf{50.97} \blue{(+4.50)}        & \textbf{70.48} \blue{(+4.66)}      & \textbf{51.63} \blue{(+6.11)}  & \textbf{70.05} \blue{(+3.83)}  & \textbf{51.30} \blue{(+5.30)}  \\  \hline
                
		\end{tabular}}
	\vspace{-0.3cm}
	\label{tab:ablation}
\end{table*}

\subsection{Analysis}
\label{sec:abla}
\smallskip
\noindent\textbf{Effect of Each Component.} 
Our method consists of several key components, including the Diversified Pixel Difference Attention module (DPDA), Multi-source Cross-attention (MSCA), Inverted Order Prediction (IOP), Shape-guided Gaussian Mapping (SGM), and Sequential Manipulation Contrastive Learning (SMCL). To evaluate the effectiveness of each component, we conduct a series of ablation studies on the facial-components track using the stem of ResNet-34 and ResNet-50. 
Specifically, 
\textbf{1)} the baseline model (BL) contains only basic encoders and decoders. 
\textbf{2)} Adding the DPDA module to the baseline model and directly feeding the features from the encoder into the sequential decoder. 
\textbf{3)} Employing the MSCA strategy for feature integration. 
\textbf{4)} Incorporating inverted order prediction. 
\textbf{5)} Inserting the SGM module into the MSCA strategy. 
{\textbf{6)} Adding the Sequential Manipulation Contrastive Learning (SMCL) loss.}
The results shown in Table~\ref{tab:ablation} indicate the contribution of each component to the overall performance. {The top part shows the performance on Facial-components dataset. With Resnet-34, the DPDA module improves the FACC and AACC by $0.62\%$ and $0.66\%$, respectively. The MSCA strategy further enhances the performance by $0.67\%$ in FACC and $1.21\%$ in AACC. The IOP significantly boosts FACC by $2.48\%$ and AACC by $4.14\%$, showing its effectiveness in gradually capturing manipulation traces. After integrating SGM, we averagely achieve 0.99\% and 1.55\% gain in FACC and AACC. By adding SMCL, the performance has a further enhancement, improving 0.4\% and 0.47\% in FACC and AACC on average.} 
The bottom part of Table~\ref{tab:ablation} corresponds to the ablation experiment on the facial-attributes dataset, showing the consistent trend that adding different components in order gradually improves the performance.

\smallskip
\noindent\textbf{Generalizability of IOP.}
The proposed IOP strategy is applicable to many architectures. To verify this, we evaluate the performance of different methods with/without IOP. As shown in Table~\ref{tab:abla-order}, all methods have gained enhancement on average. Even the methods of MA, DRN, and two-stream that are not specifically designed for sequence detection tasks surprisingly improve their performance by a notable margin. In particular, by applying IOP, Two-Stream even achieves competitive performance with MMnet on the facial-components track, demonstrating the generic effect of IOP for SDD.

\begin{table*}[!ht]
\centering
\small
\caption{\small Performance of applying IOP on various methods.}
\vspace{-0.2cm}
\resizebox{\linewidth}{!}{
\begin{tabular}{l|c|c|c|c|c|c}
\hline
\multirow{2}{*}{Methods}         & \multicolumn{2}{c|}{Facial-components}     & \multicolumn{2}{c|}{Facial-attributes}   & \multicolumn{2}{c}{Average}\\ \cline {2-7}
& FACC     & AACC      & FACC     & AACC  & FACC     & AACC  \\ \hline
Two-Stream      & 72.00     & 53.97    & 65.20     & 44.39  & 68.60     & 49.18  \\  
Two-Sream+IOP     &  73.55(\blue{+1.55})    & 58.03(\blue{+4.06})   & 67.60(\blue{+2.40)}       & 47.78(\blue{+3.39})  & 70.58(\blue{+1.98})   & 52.91(\blue{+3.73}) \\  \hline
DRN      & 69.38     & 49.76    & 67.66     & 47.31     & 68.52     & 48.54 \\  
DRN+IOP     & 71.81(\blue{+2.43})     & 54.21(\blue{+4.45})   & 68.89(\blue{+1.23})      & 49.74(\blue{+2.43})      & 70.35(\blue{+1.83})     & 51.98(\blue{+3.44}) \\     \hline
MA      & 70.12     & 51.82     & 68.91     & 49.19     & 69.52     & 50.51\\
MA+IOP  & 73.41(\blue{+3.29})     & 57.30(\blue{+5.48})     & 68.13(\red{-0.78})     & 48.58(\red{-0.61})     & 70.77(\blue{+1.25})     & 52.94(\blue{+2.43})  \\  \hline
SeqFake-Former      & 70.97     & 52.65    & 68.40     & 48.93  & 69.69     & 50.79  \\  
SeqFake-Former+IOP     & 73.34 (\blue{+2.37})     & 56.93 (\blue{+4.28})   & 69.09(\blue{+0.69})       & 48.40 (\red{-0.53})     & 71.22 (\blue{+1.53})     & 52.67(\blue{+1.88}) 
 \\  \hline
\end{tabular}
}
\label{tab:abla-order}
\vspace{-0.3cm}
\end{table*}

\smallskip
\noindent\textbf{Generalizability of SMCL.}
Note that SMCL is architecture-agnostic but specifically tailored for the SDD task, making it a potentially plug-and-play strategy. To assess its generalizability, we conduct additional experiments using different models with and without SMCL. The results in Table~\ref{tab:abla-scl} demonstrate that incorporating SMCL consistently improves performance in most cases, with average gains of approximately 0.48\% and 0.46\ in FACC and AACC, respectively. These results corroborate the efficacy of the SMCL strategy.

\begin{table*}[!ht]
\centering
\small
\caption{\small {Performance of applying SMCL on various methods.}}
\vspace{-0.2cm}
\resizebox{\linewidth}{!}{
\begin{tabular}{l|c|c|c|c|c|c}
\hline
\multirow{2}{*}{Methods}         & \multicolumn{2}{c|}{Facial-components}     & \multicolumn{2}{c|}{Facial-attributes}   & \multicolumn{2}{c}{Average}\\ \cline {2-7}
& FACC     & AACC      & FACC     & AACC  & FACC     & AACC  \\ \hline
Two-Stream      & 72.00     & 53.97   & 65.20     & 44.39  & 68.60     & 49.18 \\  
Two-Sream+SMCL     &  72.35(\blue{+0.35})    & 54.56(\blue{+0.59})   & 66.24(\blue{+1.04)}       & 45.98(\blue{+1.59})  & 69.35(\blue{+0.65})   & 50.27(\blue{+1.09}) \\  \hline
DRN      & 69.38     & 49.76    & 67.66     & 47.31     & 68.52     & 48.54 \\  
DRN+SMCL     & 69.78(\blue{+0.40})     & 49.40(\red{-0.36})   & 68.08(\blue{+0.42})      & 47.33(\blue{+0.02})      & 68.93(\blue{+0.41})     & 48.37(\red{-0.17}) \\     \hline
MA      & 70.12     & 51.82     & 68.91     & 49.19     & 69.52     & 50.51\\
MA+SMCL  & 70.72(\blue{+0.60})     & 52.38(\blue{+0.56})     & 69.21(\blue{+0.30})     & 49.57(\blue{+0.38})     & 69.97(\blue{+0.45})     & 50.98 (\blue{+0.47})  \\  \hline
SeqFake-Former      & 70.97     & 52.65    & 68.40     & 48.93  & 69.69     & 50.79  \\  
SeqFake-Former+SMCL     & 71.66 (\blue{+0.69})     & 53.35 (\blue{+0.70})   & 68.49(\blue{+0.09})       & 49.11 (\blue{+0.18})     & 70.08 (\blue{+0.39})     & 51.23(\blue{+0.44}) 
 \\  \hline
\end{tabular}
}
\label{tab:abla-scl}
\end{table*}

\smallskip
\noindent\textbf{Study on Loss Weights $\lambda_1,\lambda_2$, $\lambda_3$}. 
We study the effect of various loss weight $\lambda_{1}$, $\lambda_{2}$ and $\lambda_{3}$. As shown in Table~\ref{tab:factor}, our method demonstrates low sensitivity to these weights, with performance fluctuations limited to approximately $(\pm1.5\%)$. This is because the CE loss plays a central role in guiding the learning process and maintaining baseline performance. Despite varying $\lambda_{1}$, $\lambda_{2}$, and $\lambda_{3}$, the performance remains relatively stable. It is noteworthy that $\lambda_{3}$ has more impact than others, likely due to its direct influence on learning manipulation patterns. To strike a balanced trade-off, we set $\lambda_1=0.1,\lambda_2=0.01, \lambda_3=0.01$ in the main experiment.

\begin{table*}[!ht]
\centering
\small
    \caption{\small Different loss weights $\lambda_1,\lambda_2, \lambda_3$.}
\vspace{-0.2cm}
		\begin{tabular}{lll|c|c}
			\hline
			\multicolumn{1}{c}{$\lambda_1$}    & \multicolumn{1}{c}{$\lambda_2$} & \multicolumn{1}{c|}{$\lambda_3$} & {FACC}   &  {AACC} \\ \hline
                0.1     & 0.1    & 0.1    & 74.50 & 58.97 \\  
			
                0.1 & 0.01    & 0.1                     & 74.89 & 59.03   \\
                0.5 & 0.01   & 0.1                    & 74.69     & 58.78 \\   \hline
                \rowcolor{hl} 0.1 & 0.01  & 0.01     & \textbf{75.94}       & \textbf{60.16}    \\ \hline               
		\end{tabular}
	\vspace{-0.3cm}
	\label{tab:factor}
\end{table*}

\smallskip\noindent\textbf{Effect of Using Different Offsets in Attention Heads.}
Section ~\ref{sec:shape} introduces the shape-guided Gaussian mapping combined with multi-head cross-attention. This process generates an independent head-specific offset $\Delta \bm{\mu_{i}}$ and $\Delta \bm{\sigma_{i}}$ for each attention head. To verify the effect of offset, we remove this offset and make all attention heads share the same mean $\bm{\mu}$ and variance $\bm{\sigma}$. The results in Table \ref{tab:offset} indicate a performance drop of $0.51\%$ in FACC and $0.92\%$ in ACC without the offset, highlighting the importance of the head-specific offset.

\begin{table*}[!ht]
\centering
\small
\caption{\small Effect of using offsets in attention heads.}
\vspace{-0.2cm}
\begin{tabular}{c|c|c}

\hline
\multirow{1}{*}         & \multicolumn{1}{c|}{FACC}     & \multicolumn{1}{c}{AACC}  \\
\cline{1-3}
w/o offset  & 74.70     & 58.82 \\ \hline
w/ offset      & 75.21 (\blue{+0.51})    & 59.74 (\blue{+0.92}) \\ \hline
\end{tabular}
\vspace{-0.5cm}
\label{tab:offset}
\end{table*}

\smallskip\noindent\textbf{Effect of Adaptive Coefficient Vector $\bm{\alpha}$.}
This part studies the effect of using adaptive coefficient vector $\bm{\alpha}$ (refer to Section ~\ref{sec:dpda}). Our method adaptively balances the attention to central, angular, and radial difference operations, with $\bm{\alpha} = {\rm Softmax} ( \mathcal{H}(\mathbf{x}'))$. To demonstrate its effectiveness, we manually set fixed values $\bm{\alpha}= \{1/3,1/3,1/3\}$, equalizing the importance of the three convolutional differential operations. As shown in Table \ref{tab:adap-alpha}, the adaptive $\bm{\alpha}$ with the learnable coefficients achieves higher accuracy than the fixed weight coefficients, as expected.

\begin{table*}[!ht]
\centering
\small
\caption{\small Effect of adaptive coefficient vector $\bm{\alpha}$.}
\vspace{-0.2cm}
\begin{tabular}{c|c|c}

\hline
\multirow{1}{*}         & \multicolumn{1}{c|}{FACC}     & \multicolumn{1}{c}{AACC}  \\
\cline{1-3}
$\bm{\alpha}= \{1/3,1/3,1/3\}$      & 74.76     & 58.74 \\  \hline
$\bm{\alpha} = {\rm Softmax} ( \mathcal{H}(\mathbf{x}'))$     & 75.21 (\blue{+0.45})    & 59.74 (\blue{+1.00}) \\ \hline
\end{tabular}
\vspace{-0.4cm}
\label{tab:adap-alpha}
\end{table*}

\smallskip\noindent\textbf{Further Exploration of Prediction Order.}
This part further discusses the effect of using various prediction orders. As studied in the main text, the inverted order prediction performs better than the regular forward order. However, the performance of using mixed order has not been explored. Note that the annotation length for facial manipulation is five. Besides the complete forward order (5FO) and inverted order (5IO), we also study a set of mixed orders, which includes three forward orders and two inverted orders (3FO, 2IO), and two forward orders and three inverted orders (2FO, 3IO). The results of TSOM on the facial-components are shown in Table~\ref {tab:mixorder}, indicating that the fully inverted order performs best.

\begin{table*}[!ht]
\centering
\small
\caption{\small Effect of various orders.}
\label{tab:mixorder}
\vspace{-0.2cm}
	\begin{tabular}{c|c|c}
		\hline
		Methods & FACC & AACC \\ \hline
        5 FO, 0 IO     &   71.85 & 54.26 \\ \hline
		3 FO, 2 IO     &   71.73 & 53.59 \\ \hline
		2 FO, 3 IO     &   72.76 & 55.89 \\ \hline
		0 FO, 5 IO     &   75.21 & 59.74 \\ \hline
	\end{tabular}
\vspace{-0.2cm}
\end{table*}

\smallskip
\noindent\textbf{Adaptation on One-step DeepFake Detection (ODD).} 
Since one-step detection does not require sequential clues, existing SDD methods can not be directly applied to this task. For more comprehensive studies, we adapt these methods by replicating the one-step label $N=5$ times as the sequential annotations, where $N$ is the maximum manipulation length in dataset. For example, the real faces are assigned with annotations of $[0, 0, 0, 0, 0]$, while forged faces correspond to $[1, 1, 1, 1, 1]$. For comparison, we train these methods on the FF++ dataset~\cite{roessler2019faceforensicspp} under this configuration. Note the basenetwork of SeqFake-Former and ours is ResNet-34. The results are shown in Table~\ref{tab:FF++}, revealing that our method still achieves the best results.
We also evaluate the extended TSOM++ on the FF++ dataset. Compared to the original TSOM, TSOM++ slightly improves the performance by 0.18\%, which is less significant than the gains observed in the SDD task. This finding is consistent with our expectations, as the ODD setting does not involve sequential manipulations. Consequently, the SMCL module is reduced to a basic contrastive learning mechanism in this context, resulting in only modest performance enhancement.

\begin{table*}[!ht]
\centering
\small
    \caption{\small Performance on FF++.}
 \vspace{-0.2cm}
    \begin{tabular}{c|c}
        \hline
        \multirow{1}{*}{Methods}  &\multicolumn{1}{c}{ACC}   \\ \hline

        Two-stream                          & 82.40       \\ 
            DRN                        & 81.42    \\ 
            MA                      & 82.64           \\ 
            SeqFake-Former                  & 82.43 \\
            \rowcolor{hl} \textbf{TSOM}              & 83.70     \\ 
            \rowcolor{hl}\textbf{TSOM++}      & \textbf{83.88}  \\ \hline
           
    \end{tabular}
\label{tab:FF++}
\vspace{-0.2cm}
\end{table*}

\begin{figure}[!t]
    \centering
    \includegraphics[width=0.7\linewidth]{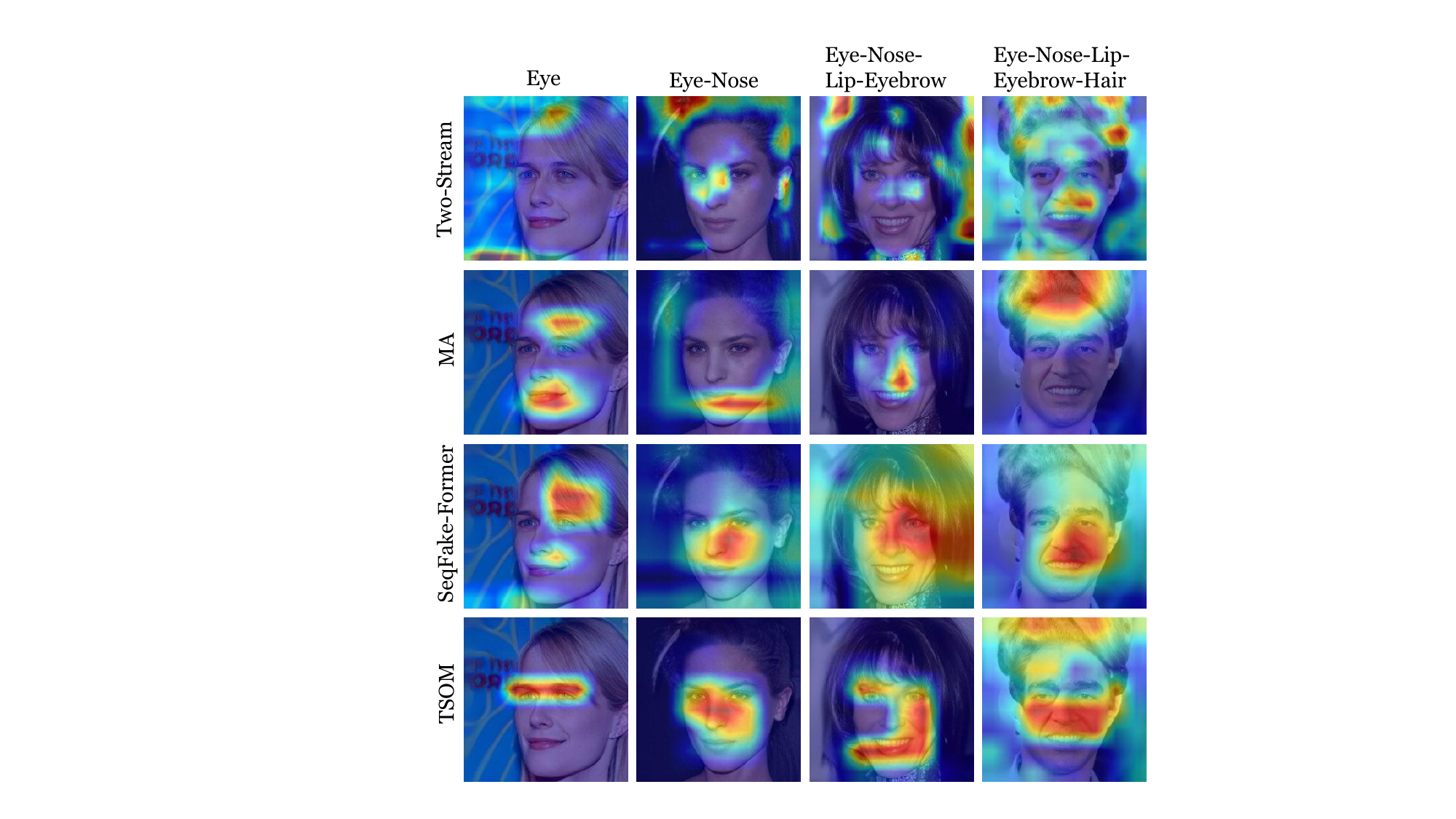}
    \vspace{-0.2cm}
    \caption{\small Attention visualization.}
    \label{fig:visualization}
\end{figure}

\smallskip
\noindent\textbf{Attention Visualization.}
Fig.~\ref{fig:visualization} shows the Grad-CAM~\cite{ramp2019cam} of different methods. We observe that under the same annotation, our method can identify the manipulated regions more precisely. In contrast, Seq-FakeFormer can localize the major manipulated regions but may overlook small manipulations with subtle traces. This result highlights the effectiveness of our method in capturing sequential manipulation traces.
Furthermore, we also observe that our method can highlight regions that are not annotated as manipulation. We attribute this to the imperfect nature of GAN in the feature space, causing subtle influences on some neighboring areas during the multi-step facial manipulation process. Our method can capture this subtle alternation as evidence for exposing the sequential manipulations.

\begin{table*}[!t]
\centering
\small
\caption{\small Performance of our methods re-evaluated regardless of manipulation orders.}
\vspace{-0.2cm}
\resizebox{\linewidth}{!}{
\begin{tabular}{l|c|c|c|c|c|c}
\hline
\multirow{2}{*}{Methods}         & \multicolumn{2}{c|}{Facial-components}     & \multicolumn{2}{c|}{Facial-attributes}   & \multicolumn{2}{c}{Average}\\ \cline {2-7}
& FACC     & AACC      & FACC     & AACC  & FACC     & AACC  \\ \hline

TSOM (ResNet-34)      & 89.90     & 79.67   & 94.09     & 85.60  & 92.00     & 82.64  \\  
TSOM++ (ResNet-34)  & 91.45     & 80.14     & 94.67     & 85.93 & 93.06
& 83.04 \\
TSOM (ResNet-50)    & 90.31      & 79.17    & 94.16       & 85.70      & 92.24      & 82.44\\  
TSOM++ (ResNet-50) & 90.98  & 79.65     & 94.78     & 86.36 & 92.88 &83.01  \\ \hline
\end{tabular}
}
\label{tab:order-ana}
\end{table*}
\smallskip
\noindent\textbf{Limitations and Future Works.}
Since current SDD methods, including ours, mainly focus on capturing sequential manipulations, they usually show limitations on the ODD problem compared to ODD-dedicated methods. Thus, one promising direction for future work is to develop an effective feature extraction component for single-step manipulation, which can be integrated to improve generalizability across both SDD and ODD tasks. Moreover, we conduct an additional tentative study, where we re-evaluate our performance by considering only whether all manipulation types in a sequence were identified, regardless of their order. As shown in Table~\ref{tab:order-ana}, performance under this relaxed criterion improves significantly. These results suggest that accurately predicting the order of manipulations remains the primary performance bottleneck. Therefore, future work will focus on developing more robust temporal correlation modeling strategies to improve order prediction accuracy.


\section{Conclusion}
This paper introduces a new Transformer design, called TSOM, for sequential DeepFake detection. Our method is inspired by three perspectives: Texture, Shape, and Order of Manipulations, leading to four key improvements. Firstly, we propose a text-aware branch featuring new Diversiform Pixel Difference Attention modules (DPDA) to capture subtle manipulation traces. Secondly, we introduce a Multi-source Cross-attention module (MSCA) to explore deep correlations between spatial and sequential features. To enhance the cross-attention further, we describe a Shape-guided Gaussian Mapping (SGM) strategy to incorporate priors on manipulation shapes. Lastly, we improve performance using a simple Inverted Order prediction (IOP) strategy, which reverses the manipulation annotation order from forward to backward. Based on TSOM, we further propose an extended method, called TSOM++, which explores the inner relations among sequential manipulations within and across faces using a newly proposed Sequential Manipulation Contrastive Learning (SMCL) scheme. Experimental results on public datasets demonstrate the superiority of our methods in sequential DeepFake detection and the comprehensive studies validate the effectiveness of each component and offer promising directions for future research.

\smallskip
\noindent\textbf{Data Availability Statements.}
The data that support the findings of this study are available in \url{https://huggingface.co/datasets/rshaojimmy/Seq-DeepFake}.

\bibliography{sn-bibliography}

\end{document}